\documentclass[cameraready]{Interspeech}

\title{TDR: Tagalog Diacritic Restoration Dataset}

\author[affiliation={1,2}, correspondingauthor, equalcontribution]{Ivan Yuri}{De Leon}
\author[affiliation={1,3}, equalcontribution]{Lanz Railey}{Fermin}
\author[affiliation={1}]{Zachary Aldridge}{Africa}
\author[affiliation={1}]{Richell Isaiah}{Flores}

\address{
    $^1$ Department of Mathematics, Ateneo de Manila University, Philippines \\
    $^2$ Globe Telecom, Philippines\\
    $^3$ Center for AI Research, Philippines
}

\email{ivan.deleon@student.ateneo.edu, ecair.lfermin@deped.gov.ph, zachary.africa@student.ateneo.edu, riflores@ateneo.edu}


\usepackage{comment}
\usepackage[most]{tcolorbox}
\usepackage{booktabs}
\usepackage{multirow}
\usepackage{xcolor}
\usepackage{colortbl}

\begin{document}

\maketitle

\begin{abstract}
Homographs are words that share identical spelling but differ in meaning and pronunciation. In Tagalog, these distinctions are disambiguated in writing through \textit{tuldik} (diacritical marks), which are rarely preserved in contemporary Tagalog text. This scarcity of diacritics limits diacritic restoration, which is useful for downstream tasks such as prosody selection for text-to-speech synthesis. To address this gap, this study introduces TDR, an annotated corpus of 40 frequent Tagalog homographs constructed from news text. It also introduces a collocation-driven retrieval method to expand the dataset and mitigate severe class imbalance among minor variants. Finally, we show that TDR supplies a capability absent from pretrained models, and that this capability can be learned by small, inexpensive models as well as by fine-tuned LLMs. TDR is publicly available on Hugging Face.\footnote{\url{https://huggingface.co/datasets/tdr-research-team/tdr-dataset}}

\end{abstract}

\section{Introduction}

Homographs are words that share the same written form but differ in meaning and, in some cases, pronunciation \cite{gottlob1999reading}. Their forms are the same when written without distinguishing marks, so the intended reading must be inferred from context \cite{gorman2018homograph}. Hence, homographs are a core challenge in word sense disambiguation (WSD) \cite{madina2024polysemy}.

In Tagalog, \textit{tuldik} (diacritics) resolve this ambiguity in writing by marking syllable stress and the presence of a final glottal stop \cite{santos2018balarila}. These cues often correspond to differences in meaning: for example, the undiacritized word \textit{puno} may refer to \textit{punò} (tree) or \textit{punô} (full), depending on context. Diacritics, thus, link a word's written form to its intended pronunciation and meaning \cite{almario2014manwal}. However, contemporary Tagalog writing rarely preserves them, especially in digital communication \cite{wells2000orthographic}. Distinct words then collapse into identical spellings, increasing ambiguity for human readers and computational models \cite{maroun2017diacritics, ljubesic2016corpus}.

The omission of \textit{tuldik} has implications for downstream language technologies, such as machine translation and text-to-speech (TTS) systems \cite{ide1998wsd, tesprasit2003context}. In TTS, an undiacritized homograph may receive a plausible but unintended reading, potentially resulting in incorrect stress or glottalization. This, consequently, leads to unnatural or unintelligible synthesized speech \cite{nicolis2021homograph}. Work on Filipino grapheme-to-phoneme (G2P) conversion likewise highlights the role of sentence-level context in determining stress \cite{marqueses2026}, suggesting that automatic diacritic restoration could serve as a useful preprocessing step for pronunciation-oriented applications.

Despite this potential, automatic diacritic restoration in Tagalog remains largely unexplored. Tagalog is a low-resource language with comparatively few large-scale annotated corpora, benchmarks, and linguistic resources \cite{cruz2021improving}, and existing Tagalog corpora generally omit diacritics \cite{dita2009building, miranda2023calamancy}, limiting their use for tasks that require fine-grained phonological distinctions. The only known dedicated effort is a small, manually annotated homograph corpus evaluated with classical classifiers \cite{flores2024diacritics}. This contrasts with languages like Arabic, which has several large-scale diacritized corpora and a long history of diacritization research \cite{diacritics_almanea2021}. Without comparable resources, progress on the development of automatic diacritic restorers for Tagalog remain hindered.

To address these limitations, we introduce the Tagalog Diacritic Restoration (TDR) dataset, a manually-annotated corpus of contemporary Tagalog news text covering 40 frequent homographs, with label reliability verified through independent re-annotation. We also propose a collocation-driven framework to expand frequency of underrepresented homograph variants. Finally, we show that training on TDR provides diacritic restoration ability that pretrained models lack, for large language models (LLMs) as well as smaller encoder and classical models.

\section{Related Work}
\label{sec:rrl}
\paragraph*{TTS and Homograph Disambiguation.} A TTS front-end must map written forms to appropriate pronunciations, and for homographs, the correct choice depends on context. Prior work has addressed this problem with supervised classifiers over lexical context \cite{gorman2018homograph}, with context-word, collocation, and part-of-speech features for Thai \cite{tesprasit2003context}, and contextual embeddings \cite{nicolis2021homograph}. Hansakunbuntheung and Thatphithakkul \cite{hansakunbuntheung2017} argue that word-pronunciation pairs alone cannot evaluate such spellings, which they call heteronyms, because the correct reading is context-dependent. Tagalog homographs distinguished by \textit{tuldik} are heteronyms in this sense, as they differ in stress and final glottal stop. Philippine-language work has addressed pronunciation through G2P models for Tagalog and other local languages \cite{aquino2019} and, more recently, stress-aware sentence-level Filipino G2P \cite{marqueses2026}. To our knowledge, these works do not provide gold diacritic labels for Tagalog homographs in context. Restoring diacritics from context offers a text-level path to this
disambiguation, and TDR provides the labeled data for it.

\paragraph*{Diacritic Restoration.} Diacritic restoration has been studied across languages using lexical and rule-based methods \cite{diacritics_neme2020, diacritics_krstev2018}, machine learning classifiers like Naive Bayes and SVMs \cite{diacritics_scannell2011, diacritics_luu2012}, and more recently, neural sequence models including BiLSTMs and transformers \cite{diacritics_stankevicius2022, diacritics_dang2020, diacritics_mubarak2019}. For Tagalog, to our knowledge, \cite{flores2024diacritics} is the only prior study specifically addressing diacritic restoration. This unpublished capstone project framed the problem as a classification task, manually annotated sentences from the Tatoeba corpus covering 21 homographs, and evaluated traditional classifiers such as random forest and logistic regression. It also measured the asymmetry ratio adopted in Section \ref{subsec:asymm}, but did not address the imbalance among homograph variants. TDR builds on this work by covering 40 homographs, including the 21 from \cite{flores2024diacritics}, alongside frequent candidates identified from a news corpus. It also provides inter-annotator agreement, uses collocation-driven retrieval to increase coverage of minor variants, and evaluates both LLMs and lightweight classifiers.


\paragraph*{Association Measures.} Association measures (AMs) quantify the tendency of words to co-occur more often than expected by chance, where a \textit{collocate} (co-occurring word) associates with a \textit{node} (target word) \cite{collocstat_brezina2018}. Different AMs capture distinct aspects of association, and their usage depends on the analytical objective \cite{collocs_gablasova2017}. Common measures include LogDice, a logarithmic coefficient which captures association strength; Mutual Information-2 (MI2), which emphasizes distinctiveness while reducing low-frequency pair bias; Delta P2 ($\Delta P_2$), directional predictiveness of the node given a collocate; and $t$-score, which shows reliability of an association \cite{collocs_brezina2020}. No single AM, however, is universally optimal since their behavior can vary with factors like syntactic structure \cite{su2024association, collocs_brezina2015}. Collocations can provide information for WSD-related tasks, as different senses tend to occur with certain associations, consistent with the ``one sense per collocation'' principle \cite{bisht2006evaluation, collocs_yarowsky1993}. This is useful for retrieval: given that a homograph's variants share one spelling in undiacritized text, collocates of a minor variant can serve as search leads for articles likely to contain it (Section~\ref{sec:collocation-expansion}).

\section{Preliminary Corpus Construction}
\label{sec:pretdr}
\textbf{Overview.} We first construct a preliminary dataset, the Pre-TDR, consisting of diacritized homographs drawn from Tagalog news text. This is based on the existing BalitaNLP corpus \cite{bunag2023balitanlp}, and we restrict our focus to homographs that are both frequently occurring and genuinely confusing to native speakers.

\subsection{Homograph Selection}
Homographs covered in this study were selected using frequency, native-speaker salience, and prior work. Frequency ensures that targets occur often enough in natural text for disambiguation to matter in practice. Salience, assessed through native-speaker judgment, prioritizes homographs that readers can readily recognize as ambiguous, since a technically homographic word that is rarely confused is less useful to include as a disambiguation target.

We first extracted the 600 most frequent word forms from BalitaNLP. Then, for each form, we checked the \textit{UP Diksiyonaryong Filipino} (UPDF) and treated it as a homograph candidate if its undiacritized spelling had multiple dictionary entries differing in diacritics. This yielded 100 candidates after applying the following exclusion principles: proper nouns (e.g., \textit{Hindi}, `an Indian language'), infrequently used base verb forms (e.g., \textit{ako}, `to take responsibility'), English-derived cognates that coincidentally share the same spelling (e.g., \textit{dating}, `date'+`-ing'), and pairs distinguished by botanical senses (UPDF entries labeled \textit{Bot.}), as they rarely occur in general text.

Native Tagalog speakers among the authors then judged the remaining candidates for salience, retaining only words that are both in common use and likely to be confused by readers. Affixed forms that share a spelling with a target word, but differ in diacritics (e.g., \textit{kabil\^ang}, `opposite' vs.\ \textit{kab\'ilang}, `including') and inflected forms (e.g., \textit{hul\'i} and \textit{hul\'ing}, both `last') were treated as separate homographs. This yielded 30 homographs.

Finally, the 30-homograph list was compared with the inventory compiled by \cite{flores2024diacritics}, which was based on native-speaker usage and interest. Eleven out of 21 homographs are already in the list, suggesting that the frequency-based selection captures a good number of homographs salient to Tagalog readers. To improve coverage and build on prior work, we added the remaining 10, yielding the final set of 40 homographs (Table \ref{tab:homographs-tdr}).

\subsection{Pre-TDR Construction}
\label{sec:corpus-construction}

The initial Pre-TDR corpus was built by manually diacritizing news articles sourced from BalitaNLP, as sufficiently large-scale diacritized Tagalog text was not available. A preliminary survey of web sources, including published journal articles \cite{hasaan2024}, essays \cite{dormiendo2024}, and social media posts \cite{almario2024}, found inconsistent use of \textit{tuldik} across authors. Some texts contained partial diacritization, while others applied it more extensively, making these sources unsuitable for aggregation as reliable gold-standard annotations.

BalitaNLP \cite{bunag2023balitanlp} contains 351,755 Tagalog news articles from eight outlets, covering domains such as crime, entertainment, and sports. News text was selected over social media-derived sources because the latter tend to exhibit non-standard spelling, code-switching, and other forms of linguistic variation that compromises consistency. The breadth and standardized conventions of news text, therefore, provide an apt source for constructing a contemporary Tagalog diacritization corpus. 

For each of the 40 homographs, 220 distinct articles were sampled, with each article containing at least one instance of the target homograph and no overlap between samples for different targets. Each sample was stored in a separate folder, resulting in 8,800 total articles. Since sample articles may naturally contain other target homographs, all occurrences of any of the 40 homographs were still annotated.

A purpose-built annotation script locates each homograph instance in a given article using case-insensitive matching, displays it with a context window to $\pm$60 characters, and asks the annotator to select the appropriate homograph variant or skip the instance. The selected variant is then written into the text and logged for traceability.

The 40 target-homograph folders were divided approximately equally into three sets, each assigned to two of the three annotators. Within each set, one author served as the primary annotator and the other as reviewer, with the reviewer checking every annotated occurrence and correcting when necessary. Roles were rotated across sets so that each annotator served both roles. All annotators are native Tagalog speakers, and the UPDF was consulted alongside native-speaker judgment for ambiguous cases. For most homographs, sentence context determines the intended variant, so the reviewer's decision was deemed final. The exception is \textit{kilal\'a} / \textit{kil\'ala}, discussed in Section \ref{subsub:kilala}, which was resolved with a fixed decision rule. The initial workflow had no independent double annotation; so, inter-annotator agreement was assessed separately (Section \ref{sub:iaa}). This process created the Pre-TDR corpus, comprising 8,800 articles, 100,087 sentences, and 38,096 diacritized homograph instances.

\subsection{Homograph Asymmetry}
\label{subsec:asymm}
Following \cite{flores2024diacritics}, we quantify the imbalance of a homograph's diacritized variants using an asymmetry ratio. Let $C_1$ be the count of its most frequent variant (hereafter, major variant) and $C_2$ the count of its second most frequent (minor variant). The ratio $r$ is given by:
\begin{equation}
    r = \frac{C_2}{C_1 + C_2},
\end{equation}
which ranges from 0 (no minor variants) to 0.5 (equally frequent variants). Third variants are excluded from the computation because of their very low counts do not materially affect the comparison between the predominant variants. Only \textit{ika} has a third variant with instances (\textit{ik\^a}, 2). 

Homographs are grouped into three categories, consistent with \cite{flores2024diacritics}: \textit{extreme} ($r< 0.1$), \textit{moderate} ($0.1 \leq r < 0.25$), and \textit{workable} ($0.25 \leq r \leq 0.5$). These cutoffs are heuristic and are adopted for comparability with prior work. Accordingly, the categories are used to describe variant imbalance and to prioritize corpus expansion, rather than as empirical thresholds for model learning.

Severe imbalances can make the minor variant more difficult to learn because relatively few training examples expose a model to the contexts in which that variant occurs. Consequently, a model trained on imbalanced data may be biased towards the major variant. Furthermore, \cite{flores2024diacritics} notes that, for text data, this imbalance cannot be address via oversampling, since duplicating existing sentences does not give new, distinct examples of the minor variant. This motivates the collection of additional authentic instances of underrepresented variants.

In the Pre-TDR corpus, 10 homographs fall under workable asymmetry (\textit{galit}, \textit{taong}, \textit{puno}, \textit{pagod}, \textit{bukas}, \textit{takot}, \textit{kilala}, \textit{tapos}, \textit{kita}, \textit{galing}), 6 under moderate asymmetry (\textit{huli}, \textit{buhay}, \textit{ika}, \textit{baka}, \textit{dating}, \textit{kabilang}), and the remaining 24 under extreme asymmetry. This skew motivated the targeted expansion described in Section~\ref{sec:collocation-expansion}, which retrieves additional articles for moderate and extreme homographs, particularly by using collocates associated with the minor variant as leads.

\section{Collocation-driven Expansion}
\label{sec:collocation-expansion}
\paragraph*{Overview.} Building on the severe class imbalance seen in Pre-TDR, we expand the corpus by retrieving additional BalitaNLP articles likely to contain the minor variants of moderately and extremely asymmetric homographs. Retrieval is primarily guided by collocates of the minor variant, with fallback strategies for variants with none. The added articles increased the proportion of minor variants enough for six homographs to move up by at least one category.

\begin{table}[h]
\centering
\caption{The 40 target homographs in the Pre-TDR and TDR corpus, grouped by asymmetry type. Asterisks indicate additions from \cite{flores2024diacritics}. Cell shading denotes improvements relative to Pre-TDR (Yellow: extreme $\rightarrow$ moderate; Green: moderate $\rightarrow$ workable; Blue: extreme $\rightarrow$ workable).}
\label{tab:homographs-tdr}
\begin{tabular}{l|l|ll}
\toprule
\textbf{Workable} & \textbf{Moderate} & \multicolumn{2}{c}{\textbf{Extreme}} \\
\midrule
bukas*                      & \cellcolor{yellow!100}abala*  & bagong    & paano      \\
galing                      & baka                           & bilang    & pala       \\
galit*                      & buhay                          & gabi      & pamahalaan \\
gamit*  \cellcolor{cyan!50}& \cellcolor{yellow!100}butas*  & kaibigan  & panalo     \\
\cellcolor{green!100}huli*  & dating                         & kasama    & pangulong  \\
ika \cellcolor{green!100}   & \cellcolor{yellow!100}huling   & kaya      & patuloy    \\
kilala*                     & kabilang                       & lalaki    & sakit      \\
kita                        &                                & lamang    & sasakyan   \\
pagod*                      &                                & madaling  & silang     \\
puno*                       &                                & nasa      & tayo       \\
takot*                      &                                &           &            \\
taong                       &                                &           &            \\
tapos                       &                                &           &            \\
\bottomrule
\end{tabular}
\end{table}

\subsection{Collocate Identification}

Using the association measures (AMs) described in Section \ref{sec:rrl}, we identified collocates of each minor variant in Pre-TDR. Each collocate was scored with the four AMs in Table \ref{tab:ams}, computed from contingency tables of co-occurrence frequencies \cite{collocstat_brezina2018}. No single AM is optimal, as argued by \cite{su2024association}. so a collocate is retained only if it met all four cutoffs. These are commonly-used values in literature, but are still arbitrary since there is no current consensus on which AM values would yield the optimal collocates \cite{collocmeas_fkih2013}. Hence, requiring all four gives a conservative collocate list that mitigates the weakness of any single AM.
\begin{table}[h]
  \caption{Association measures used for collocate identification, with
  retention cutoffs. A candidate must meet all four.}
  \label{tab:ams}
  \centering
  \small
  \begin{tabular}{@{}l p{0.5\columnwidth} l@{}}
    \toprule
    \textbf{AM} & \textbf{Description} & \textbf{Value} \\
    \midrule
    LogDice \cite{dice_rychly2008} &
      Log-scaled Dice coefficient, measures strength of association regardless of corpus size. &
      $\ge 7.0$ \\
    MI-2 \cite{MI_materna2009} &
      Bias-corrected mutual information, distinguishes exclusivity of homographs. &
      $\ge 6.0$ \\
    $\Delta P_2$ \cite{deltaP_schneider2020} &
      Metric of collocation directionality, shows how strongly a collocate predicts the node. &
      $\ge 0.005$ \\
    $t$-score \cite{tscore_xu2023} &
      Co-occurrence measure with adjusted variance, gives reliability that a collocate linkage is not due to chance.&
      $\ge 2.0$ \\
    \bottomrule
  \end{tabular}
\end{table}

Before extraction, function words (ligatures, prepositions, articles, determiners, conjunctions, and enclitics, as defined in \cite{filipino_santiago1991}) were removed, so that candidates would be content words. Target homographs included in the list, like \textit{kay\^a}, were retained. Collocates were then extracted using GraphColl in \#LancsBox X v5.0.3 \cite{software_lancsbox2024} to compute for the AMs directly, while AntConc v4.2.4 \cite{software_antconc2024} served as a secondary tool for concordance.

Extraction was run under two window settings, following the Collocation Parameter Notation of \cite{collocs_brezina2015}: a \textit{rigid} window (L1-R1) for fixed phrasal constructions and a \textit{loose} window (L5-R5) for broader thematic associations. Both are commonly-used windows and were not tuned for Tagalog. Because function words were removed beforehand, window positions count the remaining words rather than the original text. 

Table \ref{tab:collocates} lists some phrases from the rigid search that served as leads, as most variants yielded only one to three candidate collocates. 

\begin{table}[h]
\centering
\caption{Representative collocates extracted for minor-variant homographs, used as search leads for corpus expansion.}
\begin{tabular}{lll}
\toprule
\textbf{Homograph} & \textbf{Collocate} & \textbf{Completed Phrase} \\
\midrule
abála     & malaking    & malaking abála \\
abála     & mabawasan   & mabawasan [ang] abála \\
báka      & kalabaw     & kalabaw [at] báka \\
buháy     & manatiling  & manatiling buháy \\
butás     & bulsa       & butás [ang] bulsa \\
datíng    & lakas       & lakas [ng] datíng \\
kabilâng  & buhay       & kabilâng buhay \\
kilála    & ninyo       & kilála ninyo \\
lalakí    & deficit     & lalakí [ang] deficit \\
sílang    & sanggol     & sílang [na] sanggol \\
\bottomrule
\end{tabular}
\label{tab:collocates}
\end{table}

\subsection{Targeted Article Retrieval}
\label{sec:article-retrieval}
\begin{figure}[h]
    \centering
    \includegraphics[width=0.8\linewidth]{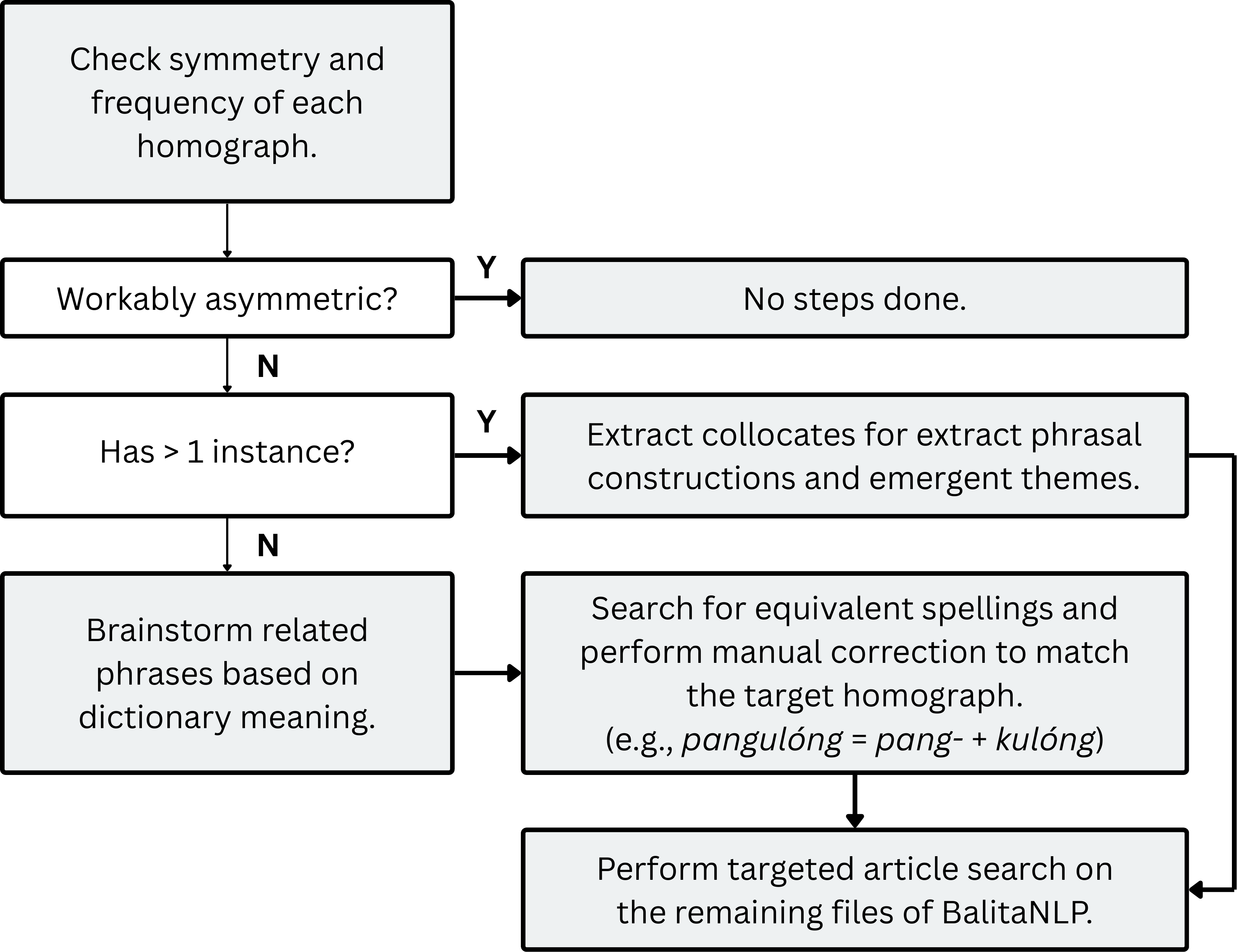}
    \caption{The decision-making process in the targeted article retrieval. Note that the brainstorming based on dictionary meaning relied on native speaker's intuition.}
    \label{fig:flowchart-TAR}
\end{figure}

Figure~\ref{fig:flowchart-TAR} describes the current work's procedure for retrieval, applied to each moderately or extremely asymmetric homograph depending how its minor variant appears in Pre-TDR. 

If the minor variant occurs more than once in Pre-TDR, rigid and loose collocation searches were done. Rigid collocates gave fixed phrases, while loose collocates were grouped together by the annotators into ``themes" and used as keyword queries even when no fixed construction exists. For instance, the loose collocates of \textit{ab\'ala} pointed to transportation disruptions and traffic.

If the minor variant never occurs, or no collocate met the thresholds defined in Table \ref{tab:ams}, two fallbacks were done: checking for alternate spellings of the homograph (e.g., \textit{pangkulong} for \textit{pangul\'ong}, and brainstorming candidate phrases and themes from the variant's meaning in UPDF (e.g., \textit{sinigang}-related articles for \textit{g\'abi}, an ingredient). Regardless of which decision path produced them, the resulting phrases and themes were queried into AntConc's Concordance tool over the remaining unannotated articles of BalitaNLP.

\section{The TDR Corpus}
\paragraph*{Overview.} This section describes the corpus produced by the target expansion described previously, its composition by homograph, and the reliability of its labels.

\subsection{Corpus Expansion}
A total of 1,265 articles were retrieved for moderately and extremely asymmetric homographs, following the process in Figure~\ref{fig:flowchart-TAR}. They were added to the corresponding  homograph folders and diacritized by the same annotator pair and procedure as Pre-TDR (see Section \ref{sec:pretdr}). 

This expansion produced the final TDR corpus, whose growth is summarized in Table \ref{tab:corpus-stats}. It contains 48,459 diacritized homographs, a 27.2\% increase from Pre-TDR. Notably, instances of non-major variants increased from 2,658 to 4,792 (80.29\%) and their share of all instances from 6.98\% to 9.89\%, suggesting that the expansion procedure worked as intended. Six homographs improved by at least one asymmetry category, with none regressing, as marked in Table~\ref{tab:homographs-tdr}. However, 101 homograph matches were not diacritized for reasons discussed in Section \ref{sec:ambiguity}.

\begin{table}[h]
\centering
\small
\caption{Corpus before and after collocation-driven expansion.}
\label{tab:corpus-stats}
\begin{tabular}{lccc}
\toprule
\textbf{Corpus} & \textbf{Articles} & \textbf{Sentences} & \textbf{Homograph Counts} \\
\midrule
Pre-TDR & 8,800  & 100,087 & 38,096 \\
TDR     & 10,065 & 146,310 & 48,459 \\
\bottomrule
\end{tabular}
\end{table}

\subsection{Homograph Breakdown}
\label{sub:iaa}
Table~\ref{tab:tdr-iaa-full} summarizes the major and minor variant counts, asymmetry ratios, and inter-annotator agreement (IAA) for all 40 homographs in the TDR corpus. After corpus expansion, 13 homographs fall under workable asymmetry, 7 under moderate, and the remaining 20 under extreme.

Most of the extremely asymmetric homographs are disproportionately high-frequency, with 11 exceeding 1,000 instances, while some are function word-like, such as \textit{kay\^a} (a connective) and \textit{l\'amang} (an adverb of degree). Despite increasing the minor variant instances, the collocation-driven procedure is possibly limited for high-frequency function words, which occur across diverse contexts rather than with distinctive collocates. Meanwhile, minor variants of eight homographs (\textit{bagong}, \textit{paano}, \textit{sakit}, \textit{panalo}, \textit{kaibigan}, \textit{madaling}, \textit{pamahalaan}, and \textit{patuloy}) remained absent after expansion.

As mentioned in Section \ref{sec:corpus-construction}, the original annotator/reviewer setup did not have independent annotation; so, an additional inter-annotator agreement (IAA) study was done to assess label reliability. A sample of 26 articles (roughly 12\% of the original 220-article pool) were reannotated as a balance between annotation effort and reliability, rather than using a formal power analysis. We sampled up to 13 articles from each variant; if the second variant had fewer than 13 instances, the remaining quota was filled by exhausting the most frequent variant until 26 articles were obtained. Homographs with fewer total instances were sampled first to avoid depleting rarer variants.

The sampled articles were independently marked by the third annotator, who had not participated in the original main annotator/reviewer pair for that homograph, and had no access to the original labels.  Disagreements were resolved by consensus with reference to UPDF.

We report free-marginal $\kappa$ \cite{randolph2005free}, an agreement coefficient for categorical labels that does not require raters to assign a fixed number of cases to each category. Chance agreement is assumed uniform across the $k$ categories, rather than from the raters' label frequencies:
\begin{equation*}
    \kappa = \frac{P_o - P_e}{1 - P_e}; \quad P_e=\frac{1}{k},
\end{equation*}
where $P_o$ is the observed agreement and $k$ is the larger of the number of variants listed for the homograph and the number observed in the sample. This ensures that $k \geq 2$ even when only a single variant appears.

Agreement in the reannotated sample was high overall, with a pooled $\kappa$ of 0.977 over 8,072 annotated instances. Note that, however, this figure is dominated by highly imbalanced homographs, where agreement is easy to reach when one variant dominates the sample. Mean $\kappa$ is slightly lower for workable (0.938) and moderate (0.95) homographs than for extreme (0.99). The lowest values are \textit{pagod} (0.848) and \textit{kabilang} (0.859). Both have substantial minor-variant representation, but the source of their lower agreement requires further examination. Nonetheless, both values exceed 0.80, which is conventionally interpreted as strong agreement \cite{landis1977measurement}. We therefore consider the labels sufficiently reliable for use, while noting that these two homographs are likely to contain residual annotation errors.

\begin{table*}[h]
\centering
\footnotesize
\caption{Variant counts, asymmetry ratio, and inter-annotator agreement (free-marginal $\kappa$) for all homographs in the TDR corpus. Rows in red have the lowest agreement value.}
\label{tab:tdr-iaa-full}
\begin{tabular}{llllll}
\toprule
\textbf{Homograph} & \textbf{Variant 1} & \textbf{Variant 2} & \textbf{Variant 3} & \textbf{Ratio} & $\kappa$ \\
\midrule
galit      & galít: 197      & gálit: 191      &            & 0.4923 & 0.939 \\
kilala     & kilalá: 254     & kilála: 238     &            & 0.4837 & 0.956 \\
puno       & punô: 170       & punò: 158       &            & 0.4817 & 0.983 \\
taong      & táong: 666      & taóng: 608      &            & 0.4772 & 0.954 \\
galing     & gáling: 368     & galíng: 280     &            & 0.4321 & 0.975 \\
huli       & hulí: 333       & húli: 232       &            & 0.4106 & 0.959 \\
\cellcolor{red!15}pagod & \cellcolor{red!15}pagód: 205 & \cellcolor{red!15}págod: 125 & \cellcolor{red!15} & \cellcolor{red!15}0.3788 & \cellcolor{red!15}0.848 \\
bukas      & bukás: 231      & búkas: 133      &            & 0.3654 & 0.947 \\
gamit      & gámit: 441      & gamít: 255      &            & 0.3664 & 0.933 \\
takot      & tákot: 235      & takót: 119      &            & 0.3362 & 0.873 \\
tapos      & tápos: 516      & tapós: 224      &            & 0.3027 & 0.959 \\
ika        & iká: 491        & ikà: 178        & ikâ: 2     & 0.2661 & 0.920 \\
kita       & kíta: 404       & kitá: 143       &            & 0.2614 & 0.950 \\
abala      & abalá: 225      & abála: 71       &            & 0.2399 & 0.970 \\
baka       & bakâ: 804       & báka: 226       &            & 0.2194 & 0.956 \\
buhay      & búhay: 1859     & buháy: 411      &            & 0.1811 & 0.950 \\
butas      & bútas: 237      & butás: 40       &            & 0.1444 & 0.964 \\
huling     & hulíng: 660     & húling: 107     &            & 0.1395 & 0.973 \\
dating     & dáting: 1700    & datíng: 221     &            & 0.1150 & 0.978 \\
\cellcolor{red!15}kabilang & \cellcolor{red!15}kabílang: 979 & \cellcolor{red!15}kabilâng: 119 & \cellcolor{red!15} & \cellcolor{red!15}0.1084 & \cellcolor{red!15}0.859 \\
kaya       & kayâ: 5308      & káya: 363       &            & 0.0640 & 0.984 \\
lamang     & lámang: 2968    & lamáng: 150     &            & 0.0481 & 0.975 \\
silang     & siláng: 1515    & sílang: 67      &            & 0.0424 & 0.989 \\
bilang     & bílang: 3569    & biláng: 102     &            & 0.0278 & 1.000 \\
lalaki     & laláki: 737     & lalakí: 12      &            & 0.0160 & 1.000 \\
sasakyan   & sasakyán: 696   & sasákyan: 6     &            & 0.0085 & 0.988 \\
pala       & palá: 833       & pála: 1         &            & 0.0012 & 0.989 \\
tayo       & táyo: 1702      & tayô: 4         &            & 0.0023 & 1.000 \\
gabi       & gabí: 1191      & gábi: 2         &            & 0.0017 & 0.989 \\
pangulong  & pangúlong: 1386 & pangulóng: 1    &            & 0.0007 & 0.992 \\
pamahalaan & pámahalaán: 819 & pámahaláan: 0   &            & 0.0000 & 1.000 \\
patuloy    & patúloy: 1062   & patulóy: 0      &            & 0.0000 & 1.000 \\
kasama     & kasáma: 2261    & kasamâ: 2       & kasamá: 0  & 0.0009 & 0.995 \\
nasa       & nása: 3365      & nasà: 1         &            & 0.0003 & 1.000 \\
bagong     & bágong: 2039    & bagóng: 0       &            & 0.0000 & 0.992 \\
paano      & paáno: 769      & paanó: 0        &            & 0.0000 & 1.000 \\
sakit      & sakít: 606      & sákit: 0        &            & 0.0000 & 1.000 \\
panalo     & panálo: 614     & panaló: 0       &            & 0.0000 & 1.000 \\
kaibigan   & kaibígan: 698   & kaibigán: 0     &            & 0.0000 & 1.000 \\
madaling   & madalîng: 554   & madáling: 0     &            & 0.0000 & 1.000 \\
\bottomrule
\end{tabular}
\end{table*}

\subsection{Annotation Decisions}
\label{sec:ambiguity}
Beyond choosing a variant, two decisions needed explicit rules: when to leave a matched string undiacritized, and how to label \textit{kilal\'a} / \textit{kil\'ala}, where context alone did not settle the variant.

\subsubsection{Undiacritized Instances}
A total of 101 homograph-pattern matches (0.21\% of all matches in the released TDR corpus) were left undiacritized, for the following reasons.
\begin{itemize}
    \item \textbf{Misspellings.} A typo may match a target homograph. For example, \textit{``Masustansiya ang sinigang dahil marami itong sangkap na gulay \textcolor{blue}{\textbf{kaya}} ng kangkong, sitaw, okra, labanos, at talong''} (``Sinigang is nutritious because it has many vegetable ingredients like kangkong, string beans, okra, radish, and eggplant'') uses \textit{kaya} where \textit{gaya} (``like'') was intended.

    \item \textbf{Subword fragments.} A match may be part of a larger word, as in \textit{``Kailangang aregluhin pa ito ng \textcolor{blue}{magka-\textbf{bilang}} gobyerno''} (``This still needs to be settled by both governments''), where \textit{bilang} is part of \textit{magka-bilang} (``both''), or of a URL slug, e.g., \textit{.../pinoy-nagdala-ng-\textcolor{blue}{\textbf{bagong}}-covid-variant-sa-.../}.

    \item \textbf{Contracted forms.} A match may be part of a contraction. For example, \textit{``Kung noon kapag \textcolor{blue}{\textbf{ika}'y} nababagot at sa palagay mo ay kailangan mo...''} (``Back then, when you got bored and felt that you needed...'') contains \textit{ika} only as part of \textit{ika'y} (\textit{ikaw} + \textit{ay}).

    \item \textbf{Proper nouns with unknown pronunciation.} For example, \textit{``...isang torneong pinamayagpagan ni Japanese \textcolor{blue}{\textbf{Nasa}} Hataoka...''} (``...a tournament dominated by Japan's Nasa Hataoka...'') uses \textit{Nasa} as a personal name. Proper nouns with well-known pronunciations, such as \textit{Siláng, Cavite}, were diacritized, since explicit diacritics benefit TTS when the pronunciation is known.

    \item \textbf{Code-switched foreign words.} For example, \textit{``Ang hindi ko ide-deny, I'm \textcolor{blue}{\textbf{dating}} now at hindi siya ang rason...''} (``What I won't deny is that I'm dating now, and they aren't the reason...'') uses the English gerund \textit{dating}, not the Tagalog homograph \textit{d\'ating} / \textit{dat\'ing}.
\end{itemize}

\subsubsection{Disambiguating \textit{kilal\'a} and \textit{kil\'ala}}
\label{subsub:kilala}

Unlike most homographs, context alone could not reliably separate these two, and consultation with two Filipino linguistics professors gave conflicting recommendations. To ensure consistent annotation, we adapted a distinction based on the UPDF: \textit{kilal\'a} is a synonym of \textit{bantog} (famous, well-known), while \textit{kil\'ala} denotes personal acquaintanceship (cf. \textit{kakil\'ala}). The operational rule, then, bases on whether an explicit \textcolor{blue}{\textit{knower}}, the entity doing the knowing or has knowledge of a \textcolor{orange!60!black}{\textit{known party}}, appears in the sentence. If so, the instance is \textit{kil\'ala}; otherwise, it is \textit{kilal\'a}.

\begin{quote}
\textit{Kung napansin mo, parang nahihiya siya sa akin kasi nga, kil\'ala \textcolor{blue}{ko} \textcolor{orange!60!black}{siya} noon pa.} (If you noticed, they seem embarrassed around me because I've known them for a while now.)
\end{quote}
Here, the explicit mention of a \textcolor{blue}{knower} (\textcolor{blue}{\textit{ko}}, ``I'') identifies the appropriate annotation as \textit{kilála}. In contrast, \textit{kilal\'a} describes the \textcolor{orange!60!black}{known party} as being known or famous without introducing a specific knower.

\begin{quote}
\textit{Si \textcolor{orange!60!black}{De Lima} ay kilal\'a bilang isa sa mga kritiko ni Duterte.} (De Lima is known as one of Duterte's critics.)
\end{quote}
Only the \textcolor{orange!60!black}{known party} (\textcolor{orange!60!black}{\textit{De Lima}}) is expressed, with no explicit \textcolor{blue}{knower}, yielding \textit{kilalá}. A more challenging case arises when an explicit \textcolor{blue}{knower} is present while the \textcolor{orange!60!black}{known party} is also independently famous, seemingly satisfying both definitions:

\begin{quote}
\textit{Kil\'ala \textcolor{blue}{natin} si \textcolor{orange!60!black}{Ai-ai delas Alas} bilang Comedy Queen, ang BFF ni Sharon Cuneta.} (We know Ai-ai delas Alas as the Comedy Queen, Sharon Cuneta's best friend.)
\end{quote}
Ai-ai delas Alas is widely known which, in isolation, would suggest \textit{kilal\'a}. However, the annotation rule is structured to give precedence to the presence of an explicit \textcolor{blue}{knower}; in this sentence, the \textcolor{blue}{knower} is \textit{natin} (``we''). So, the instance is annotated as \textit{kil\'ala}.

This rule was applied consistently throughout the corpus to ensure annotation consistency. Nevertheless, the disagreement observed among both annotators and consulted experts suggests that the distinction between \textit{kilal\'a} and \textit{kil\'ala} is not categorically resolved in contemporary Tagalog usage. Hence, this rule should be interpreted only as an operational annotation guideline, not a definitive linguistic analysis.

\section{Experiments}
\label{sec:experiments}
We evaluate the utility of TDR on two tasks that differ in how much of the diacritic restoration problem a model must solve. \textbf{Multiple-choice disambiguation} (MCD) gives the candidate set and asks only which variant a homograph takes in context, isolating disambiguation from text generation and places classifiers, encoders, and LLMs on equal footing. \textbf{Sentence reconstruction}, meanwhile, asks a model to restore diacritics in an entire sentence -- which is closer to how restoration could serve a TTS front-end and is the natural output for LLMs. We additionally test whether reconstruction output improves a downstream TTS system through a small-scale test.

\subsection{Data Split}
\label{subsec:data-split}
TDR was partitioned, at the article level, into two disjoint sets: a test pool of 1,040 articles, corresponding to the reannotated dataset used for IAA, and a training pool of the remaining 9,025 articles. Articles were then segmented into sentences. For the test benchmark, only sentences containing at least one diacritized target homograph were retained. Sentences where the same homograph appears more than once with different variants were excluded, since these cannot be assigned a single intended variant. For instance, ``\textit{\textbf{Kay\^a} nga nagtitiyaga sila ... ito na lang ang \textbf{k\'aya} nilang bilhin.}'' (``That's precisely why they persevere ... because it's all they can afford to buy.'') contains two different variants of \textit{kaya}, and is therefore excluded.

For fine-tuning the neural models, the training pool was split approximately 90/10 at the sentence level, guided by homograph--variant groups. Separate splits yielded 28,742 training and 3,199 validation records for reconstruction, and 34,869 training and 3,882 validation questions for MCD. The test benchmark covers all 40 homographs, with 7,718 MCD items targeting distinct homographs within sentences and 6,113 reconstruction items, one per distinct undiacritized sentence.

\subsection{Models}
\label{subsec:models}
We compare non-neural baselines, fine-tuned encoders, and LLMs. No model uses class reweighting, so that each model sees the natural imbalance among homograph variants.

\begin{itemize}
    \item \textbf{Baselines.} Two baselines were evaluated. First, a majority baseline model predicts each homograph's most frequent variant. Second, a logistic regression (LR) baseline fits one classifier per homograph on TF-IDF features of the undiacritized sentence, with a maximum of 5,000 features. For each homograph, we select $C \in \{0.1, 0.3, 1, 3, 10\}$, n-gram range (unigram or bigram), and TF-IDF weighting (standard, binary, or sublinear) by stratified 5-fold cross-validation on macro-F1. The LR model is trained only on 24 homographs whose minor variant has at least five training sentences. The other 16 fall back to the majority prediction.

    \item \textbf{Encoders.} We fully fine-tune four encoders: RoBERTa-base and RoBERTa-large \cite{liu2019roberta}, pretrained mainly on English, and their Tagalog counterparts RoBERTa-Tagalog-base and RoBERTa-Tagalog-large \cite{cruz2021improving}. Each candidate is paired with the undiacritized sentence and scored by a shared head, trained with cross-entropy to rank the gold variant highest (3 epochs, learning rate $2\times10^{-5}$, effective batch size 16). The checkpoint with the best validation loss is saved.

    \item \textbf{LLMs.} We use Llama-3.1-8B-Instruct \cite{grattafiori2024llama3herdmodels} and Llama-SEA-LION-v2-8B \cite{ng2025sealionsoutheastasianlanguages}, the latter adapted for Southeast Asian languages. Each is evaluated zero-shot and fine-tuned, with a separate QLoRA adapter per task trained on that task's prompt format (4-bit NF4; rank 16, $\alpha=32$, dropout 0.05 on all attention and MLP projections; 3 epochs, learning rate $2\times10^{-4}$, effective batch size 16; loss on the answer only; best checkpoint by validation loss), giving four fine-tuned models.

\end{itemize}

Note that RoBERTa is evaluated on multiple-choice disambiguation only; LR and majority are evaluated on multiple-choice and simulated for reconstruction (see Section~\ref{subsec:recon}); while LLMs are evaluated on both.

\subsection{Multiple-Choice Disambiguation}
\label{subsec:mcq}

\paragraph*{Setup.} The model chooses the form a target homograph takes in an undiacritized sentence. Each item corresponds to one homograph instance and offers four candidate forms: all attested homograph variants, its undiacritized spelling, and invalid diacritized forms (when options are fewer than four). Candidates are shuffled once per item. LLMs receive the prompt below and answer with the chosen form. Its outputs are then NFC-normalized and parsed: an output counts as an answer if exactly one candidate appears in it (case-insensitive). Meanwhile, RoBERTa encoders score each candidate paired with the sentence and return the highest-scoring one. Baseline models predict a variant directly, so they always return a valid candidate.


\begin{quote}
\noindent\fbox{
\begin{minipage}{\linewidth}
\small
\textbf{System:} You are a Tagalog diacritic restoration assistant. From the list of variants given, respond with ONLY the correct variant, copied exactly as written (including its diacritic marks). Do not include any explanation, reasoning, punctuation, or extra text.

\vspace{3pt}
\textbf{User:} How should the word \{homograph\} be written with the correct diacritic marks in the sentence ``\{sentence\}''? \\
Options: \\
- \{choice$_1$\}\\
- \{choice$_2$\}\\
- \{choice$_3$\}\\
- \{choice$_4$\}\\
Answer:
\end{minipage}
}
\end{quote}

Since accuracy is dominated by majority variants, we use macro-F1 as the primary metric. For a homograph $h$ with variant set $V_h$ attested in its instances, we first average the per-variant F1 scores within each homograph, then average across homographs:
\begin{align}
    \text{F1}_h &= \frac{1}{|V_h|} \sum_{v \in V_h} \text{F1}_{h,v}, \\
    \textrm{Macro-F1} &= \frac{1}{|H|} \sum_{h \in H} \text{F1}_h,
\end{align}
where $H$ is the set of all homographs. Because both averages are unweighted, a minor variant counts as much as a major one, and every homograph counts equally regardless of its frequency. The latter makes the metric noisy: if a homograph's minor variant has only one or two test instances, a single error can cut that homograph's F1 by up to half. We therefore also report macro-F1 over the 20 workably and moderately asymmetric homographs (W+M F1), and weighted F1, in which each homograph's F1 is weighted by its number of test instances.

\begin{table}[h]
\centering
\footnotesize
\caption{Multiple-choice disambiguation results. W+M F1 considers workably and moderately asymmetric homographs only. Weighted: per-homograph F1 weighted by test instances.}
\label{tab:mcq-results}
\resizebox{\columnwidth}{!}{%
\begin{tabular}{lccc}
\toprule
\textbf{Model} & \textbf{Macro-F1} & \textbf{W+M F1} & \textbf{Weighted F1} \\
\midrule
Majority                  & 0.558 & 0.396 & 0.568 \\
Logistic regression   & 0.800 & 0.817 & 0.794 \\
\midrule
RoBERTa-base              & 0.654 & 0.601 & 0.646 \\
RoBERTa-large             & 0.559 & 0.505 & 0.558 \\
RoBERTa-Tagalog-base      & 0.786 & 0.817 & 0.754 \\
RoBERTa-Tagalog-large     & 0.815 & 0.844 & 0.826 \\
\midrule
Llama, zero-shot          & 0.276 & 0.307 & 0.263 \\
Llama, fine-tuned         & 0.831 & 0.878 & 0.833 \\
SEA-LION, zero-shot       & 0.330 & 0.330 & 0.301 \\
SEA-LION, fine-tuned      & \textbf{0.838} & \textbf{0.898} & \textbf{0.849} \\
\bottomrule
\end{tabular}
}
\end{table}

\paragraph*{Results.} Table \ref{tab:mcq-results} reports the results. Fine-tuned SEA-LION performs best, closely followed by fine-tuned Llama. Both improve sharply over their zero-shot counterparts, which fall below even the majority baseline. This suggests that Tagalog diacritic disambiguation is not readily available from general-purpose pretraining, and that TDR can serve as an effective fine-tuning data for acquiring it. 

TDR also supports competitive performance from much smaller models. For instance, RoBERTa-Tagalog-large comes within 0.024 macro-F1 of the fine-tuned LLMs, while LR remains competitive. Among the encoders, both Tagalog models outperform their English counterparts, with English RoBERTa-large notably trailing RoBERTa-base, suggesting that language-specific pretraining matters more than model size in this setting. The gains from fine-tuning are largest on W+M F1, where minor variants are better represented. The fine-tuned LLMs also substantially outperform the majority baseline on this subset (0.878--0.898 versus 0.396).

\paragraph*{Error analysis.}
\label{subsec:mcq-errors}
The fine-tuned LLMs rarely answered in the expected format. Over 96\% of their outputs required parsing, and fewer than 0.4\% consisted of the correct form alone. Instead of a single form, they typically repeated their chosen form, either across lines or concatenated (e.g., the word \textit{bil\'ang} written several times). Parsing recovers the intended answer in most of these cases, since only one candidate appears. About 3\% of outputs, however, contained no candidate or more than one, and were scored as incorrect. The zero-shot models show the opposite pattern: their outputs almost never required parsing, but they often chose the wrong form. Fine-tuning thus improved disambiguation substantially while weakening adherence to the answer format. A finer-grained analysis of errors by diacritic type (stress vs. glottal stop) and by position within the word is left to future work.

\subsection{Sentence Reconstruction}
\label{subsec:recon}

\paragraph*{Setup.} Each item is an undiacritized test sentence, which the model must return with diacritics restored. Unlike multiple-choice disambiguation, this task provides no candidate list. The model must identify the target homographs and choose their variants. It therefore more closely resembles restoration for a TTS front-end.

LLMs receive the prompt below. Encoders are not evaluated because they only score given candidates. For the majority-class baseline and LR, we simulate reconstruction by substituting their multiple-choice predictions for the target homographs in each sentence.

\begin{quote}
\noindent\fbox{
\begin{minipage}{\linewidth}
\small
\textbf{System:} Modify the homographs in this Tagalog sentence by adding the correct diacritic marks (tuldik).

\vspace{3pt}
\textbf{User:} \{sentence\}
\end{minipage}
}
\end{quote}

We report two sentence-level metrics. \textbf{Exact match} requires the output to equal the gold sentence. \textbf{Recovery} requires only that, for every target homograph, the variants appearing in the output match the gold variants in the same order, so changes elsewhere in the sentence are ignored. Target occurrences in the output are located by diacritic-stripped string matching. For the simulated baselines, the two metrics coincide by construction, since only target homographs are changed.

\begin{table}[h]
\centering
\caption{Sentence reconstruction results, shown in percentages. For simulated baselines ($^\dagger$), exact match equals recovery.}
\label{tab:recon-results}
\begin{tabular}{lcc}
\toprule
\textbf{Model} & \textbf{Exact Match} & \textbf{Recovery} \\
\midrule
Majority$^\dagger$        & 85.20 & 85.20 \\
LR$^\dagger$              & 93.73 & 93.73 \\
\midrule
Llama, zero-shot          & 0.00  & 0.02 \\
Llama, fine-tuned         & \textbf{95.50} & \textbf{97.04} \\
SEA-LION, zero-shot       & 0.18  & 0.74 \\
SEA-LION, fine-tuned      & 95.47 & 96.99 \\
\bottomrule
\end{tabular}
\end{table}
\paragraph*{Results.} Table~\ref{tab:recon-results} shows the results. Zero-shot LLMs almost never return a correctly restored sentence (at most 0.74\% recovery), indicating that its diacritic restoring capability is absent without task-specific training. After fine-tuning on TDR, Llama and SEA-LION are practically tied as they recover the gold variants in about 97\% of sentences, and match the gold sentence in about 95.5\%. The 1.5-point gap between recovery and exact match rates reflect occasional edits done outside the target homographs which include, but are not limited to, whitespace, punctuation, or changes to other words. 

The majority baseline already recovers 85\% of sentences, since most contain only major variants, so any further gain must come from the remaining sentences that require a minor variant. On both metrics, the fine-tuned LLMs outperform the zero-shot LLMs as well as the simulated majority and LR baselines. Because these baselines are given the target positions, this comparison is conservative with respect to the LLMs.

\subsection{Small-scale TTS check}
\paragraph*{Setup.} We test whether restored diacritics improve pronunciation using MOSS-TTS-v1.5, a Tagalog-capable TTS model \cite{gong2026mossttstechnicalreport}. Each item is synthesized in three conditions: undiacritized, LLM-restored, and gold (diacritized). Five sentences were sampled for the major and minor variants of the 13 workably asymmetric homographs and synthesized using the same seed and voice. For LLM-restored inputs, the fine-tuned Llama restores the sentence, after which only its prediction for the target homograph is spliced back to preserve matched contexts. Three independent annotators identify the variant pronounced in each clip or mark it unintelligible. The majority vote determines the perceived variant, and homograph pronunciation error rate (HPER) is the proportion of clips whose perceived variant differs from the gold variant.

\begin{table}[h]
\centering
\caption{HPER by input text, based on the majority vote of annotators. Clips with invalid LLM output and marked as unintelligible are excluded.}
\label{tab:tts}
\begin{tabular}{lcc}
\toprule
\textbf{Input Text} & $N$ & \textbf{HPER} \\
\midrule
Undiacritized        & 129 & 35.66 \\
LLM-restored         & 129 & \textbf{12.40} \\
Gold (diacritized)   & 130  & 11.54 \\
\bottomrule
\end{tabular}
\end{table}

\paragraph*{Results.} Table \ref{tab:tts} shows that feeding the TTS model text restored by the TDR-finetuned model reduced the HPER from 35.66\% to 12.40\%, gaining nearly the same improvement obtained with the actual diacritized sentences from TDR. This corresponds to a relative reduction of about 65\%, leaving the LLM-restored input nearly identical to the HPER of the gold diacritization. Inter-annotator aggrement was moderate (free-marginal $\kappa = \text{0.52}$), which may partly reflect annotators occasionally choosing the actual sentence context rather than the audio. This motivates stricter listening protocols in future work. 

Although small in scale, these results suggest that fine-tuning diacritic restoration models with TDR can be a useful step for improving TTS pronunciation of Tagalog words, especially homographs, whose pronunciation cannot be determined from spelling alone.

\section{Limitations}
TDR covers 40 homographs drawn from a single domain, news text from BalitaNLP, so findings may not transfer to other registers. Independent reannotation covered only the 1,040-article test pool. Collocation-driven expansion did not fully resolve class imbalance: function word-like homographs remain extremely asymmetric, and eight still had no minor variant. Because some minor variants have only one or two test instances, per-homograph F1 on extremely asymmetric homographs is unstable; W+M F1 mitigates but does not remove this. The benchmark also excludes sentences in which one homograph appears with conflicting variants, which a practical restorer must handle.

On the modeling side, we evaluate two 8B LLMs and four encoders, each fine-tuned in a single run with fixed hyperparameters, and report no confidence intervals. Fine-tuned LLMs required output parsing in multiple choice, so their scores depend on the parsing rule. A TTS-relevant error analysis, examining which restoration errors most affect synthesized pronunciation, is recommended for future work. Finally, the TTS check only used a single system and sentences from workable homographs. 

\section{Conclusion}
We introduced TDR, a manually annotated corpus of 40 Tagalog homographs with diacritic marks in context, comprising 48,459 diacritized instances from 10,065 news articles. Collocation-driven retrieval increased the number of minor-variant instances by about 80\%, and independent reannotation of the test pool showed high agreement ($\kappa = 0.977$). The evaluated zero-shot LLMs fell below the majority baseline in multiple-choice disambiguation and almost never restored a sentence correctly. Fine-tuning on TDR yielded the best performance among the evaluated models on both tasks, recovering the gold variants in about 97\% of test sentences. Smaller models were also competitive: RoBERTa-Tagalog-large approached the macro-F1 of the fine-tuned LLMs, while per-homograph logistic regression trailed them by only a few points. TDR is publicly available on Hugging Face under the license conditions of BalitaNLP, supporting further work on Tagalog diacritic restoration and pronunciation-aware speech technology.

\section{Generative AI Use Disclosure}
Generative AI tools were used to assist with editing prose for clarity and conciseness, and with drafting portions of code used in data processing and experimentation. All research design, data annotation, experimental results, and analytical conclusions are the authors' own. AI-assisted content was reviewed and verified by the authors prior to inclusion.

\bibliographystyle{IEEEtran}
\bibliography{mybib}

\end{document}